\documentclass[
hf,
]{ceurart}

\usepackage[vlined, linesnumbered ]{algorithm2e}
\usepackage{amsmath,amssymb,amsfonts}
\SetKwInput{KwInput}{Input}                % Set the Input
\usepackage{amsmath,amsfonts,amssymb,mathrsfs}
\begin{document}

% \copyrightyear{2021}
% \copyrightclause{Copyright for this paper by its authors.
%   Use permitted under Creative Commons License Attribution 4.0
%   International (CC BY 4.0).}

\conference{ArXiv preprint}

\title{Automatic Visual Inspection of Rare Defects: A Framework based on GP-WGAN and Enhanced Faster R-CNN}

\author[1,2]{Masoud Jalayer}[%
orcid=0000-0001-8013-8613,
email=masoud.jalayer@polimi.it,
]

\author[3]{Reza Jalayer}[%
email=reza.jalayer74@gmail.com,
]

\author[4]{Amin Kaboli}[%
email=amin.kaboli@epfl.ch,
]

\author[1]{Carlotta Orsenigo}[%
email=carlotta.orsenigo@polimi.it,
]

\author[1]{Carlo Vercellis}[%
email=carlo.vercellis@polimi.it,
]

\address[1]{Department of Management, Economics and Industrial Engineering, Politecnico di Milano, Via Lambruschini 24/b, 20156, Milan, Italy}

\address[2]{Advanced Control \& Intelligent Systems Laboratory, School of Engineering, University of British Columbia, Kelowna, BC, Canada}

\address[3]{Center of Excellence in Experimental Solid Mechanics and Dynamics, School of Mechanical Engineering, Iran University of Science and Technology, Narmak 16846, Tehran, Iran}

\address[4]{Institute of Mechanical Engineering, School of Engineering, Swiss Federal Institute of Technology at Lausanne (EPFL), Lausanne, Switzerland}

\maketitle
\begin{abstract}
A current trend in industries such as semiconductors and foundry is to shift their visual inspection processes to Automatic Visual Inspection (AVI) systems, to reduce their costs, mistakes, and dependency on human experts. This paper proposes a two-staged fault diagnosis framework for AVI systems. In the first stage, a generation model is designed to synthesize new samples based on real samples. The proposed augmentation algorithm extracts objects from the real samples and blends them randomly, to generate new samples and enhance the performance of the image processor. In the second stage, an improved deep learning architecture based on Faster R-CNN, Feature Pyramid Network (FPN), and a Residual Network is proposed to perform object detection on the enhanced dataset. The performance of the algorithm is validated and evaluated on two multi-class datasets. The experimental results performed over a range of imbalance severities demonstrate the superiority of the proposed framework compared to other solutions.
\end{abstract}



\section{Introduction}
\subsection{Motivation}
The paper proposes a defect detection framework towards an efficient Automatic Visual Inspection (AVI) system for industries that lack sufficient amount of annotated samples. Nowadays, due to their high generalization abilities and easy-to-use nature, AVI systems based on deep learning (DL) models have been proposed everywhere in the literature. Nonetheless, DL-based classifiers require an adequate amount of samples to reach high performances. Furthermore, creating a large and accurately labelled dataset needs a high number of defective samples and a considerable amount of time for the annotation. On the other hand, annotation is the requirement of any image object detection algorithm. In this paper, therefore, alongside an enhanced image object detection model for AVI systems, we propose a novel data augmentation model that alleviates this burden for the practitioners and automatically synthesizes and annotates new images from a limited number of samples.
\subsection{AVI systems}
AVI plays a key role in quality control and process monitoring in the automated production line technologies. It is aimed to ensure that the product is healthy based on a visual control and there is no defect on its surface. A successful implementation of AVI can drastically lower the production costs and respond to the increasing demand of high-quality and reliable productions in different industries. Hence, in recent years, AVI systems have been receiving more and more research attention in quality control and, as a comprehensive technology, they have been widely used in industries such as steel and aluminum productions \cite{Zhang2020a}, electronic devices and boards \cite{Zheng2021a}, railway health monitoring \cite{Niu2021}, wood flooring \cite{Delgado}, silicon wafers \cite{Yuan-Fu2020}, fabric and textile materials \cite{Tsang2016a} and fruit and vegetable productions \cite{Cubero2011}.\\
AVI systems are chiefly configured by two parts, represented by optical imaging devices and an image processing software. These systems can comprise different types of cameras to detect moving products and their components, and can be regarded as decision support systems able to conduct quality control based on the images they receive in input. Fundamentally, the image processing software of an AVI method accomplishes two main tasks: (1) feature extraction, which contains a set of hand-crafted and automatic encoding-decoding techniques to pull out the defect signs; (2) classification, which incorporates a set of methods to diagnose the sample condition based on the extracted features.\\
Current AVI image processing systems are mostly based on hand-crafted feature extraction techniques, such as threshold models, statistical methods, filter-based techniques, transform-based and model-based methods, requiring skilled workers with prior knowledge of the specific application and domain. Such systems need long development time and more complex installations, and lack the generalization ability required to be easily applied to other similar settings and applications.
It is worthwhile to observe that, state-of-the-art AVI systems, based on deep learning (DL) architectures, do not require expert designers and have a satisfactory generalization ability. However, they usually get poor results when they are dealing with insufficient number of samples and imbalanced datasets, as in the case of real-world applications, in which some defect types happen infrequently. This hinders current AVI systems to perform accurately.

\subsection{Related works}
Some AVI methods are characterized by spatial distribution of grayscale images: Tsai et al. \cite{Tsai2019} proposed a mathematical morphology technique to detect the machine circular tool-mark textures. \cite{Merazi-Meksen2014} introduced a similar mathematical model to extract relevant pixels corresponding to the presence of a discontinuity for crack recognition. Wu and Hu \cite{Wu2003} suggested co-occurrence matrices of metal grayscale images to access the information of metal surfaces. Despite their quick runtime, these methods are highly dependent on the interference conditions and presence of noise. Moreover, they are not capable of detecting rare patterns. Traditional machine learning techniques, such as Support Vector Machines \cite{Wu2019}, Decision Trees\cite{Aghdam2011}, Naive Bayesian network \cite{Pernkopf2004}, are also widely used in the literature of AVI systems. However, these so-called ``shallow learning''-based systems have shown shortcomings when they are dealing with images of multi defect types, noisy samples or rare patterns\cite{Jalayer2021}.

In the literature of AVI systems an increasing attention has been paid to DL-based computer vision algorithms. Convolutional neural networks (CNN) have boosted the image classification performance and have been employed in different quality control systems. Soukup et al. \cite{Soukup2014} introduced a CNN-based setup for rail surface inspection using photometric stereo images. Weimer et al. \cite{Weimer2016} investigated different CNN hyper-parameters involved in the process of an industrial optical inspection. To discriminate defect-free and defective samples, \cite{Tsai2021} employed a convolutional auto-encoder (CAE) architecture with two different regularization penalties which confine the spread of the extracted features in a very limited space from a set of defect-free samples and divide the background texture. These DL-based models are mostly efficient when they cope with image classification problems. On the other hand, to locate the defects in the sample images, image object detection and image segmentation techniques should be used. Enshaei et al. \cite{Enshaei2020} proposed a U-Net image segmentation architecture for different surface types. \cite{Hao2020a} developed an AVI system of steel surfaces based on Fast R-CNN architecture with a feature fusion network, generating high-quality multi-resolution feature maps for the inspection of multi-size defects. In comparison with R-CNN and Fast R-CNN, Faster R-CNN can achieve a higher accuracy with a considerable faster training speed. In terms of runtime, Faster R-CNN is almost ten times as quick as Fast R-CNN, which makes it suitable for on-line detection applications. Compared to some one-step detectors such as YOLO-v3, Faster R-CNN is slightly slower. However, it achieves more accurate solutions in different domains and image contexts\cite{Liu2021}. \cite{Shi2017} used a Faster R-CNN to detect and locate the fiber paper tube defects. To detect the printed circuit boards, \cite{Hu2020} also employed a Faster R-CNN architecture, coupled with a ResNet-50 for its feature extraction ability. In spite of their satisfactory performances, these solutions require a sufficient number of samples for each defect type, which is in practice hard to obtain. DL-based solutions need a huge amount of samples representing different defect types and defect-free conditions. To respond to this need, Niu et al. \cite{Niu2020} proposed a generative algorithm called surface defect-generation adversarial network (SDGAN) which creates synthetic samples suitable for image classification, since it does not generate image annotations but only image labels.

\subsection{The Key Contributions}
The main contributions of this paper are four-fold:
\begin{itemize}
    \item We propose a new automated visual inspection system combining a generative algorithm and an image object detection model aimed to conduct the automated surface inspection on various products and to alleviate the adverse effect of imbalanced datasets.
    \item We propose a heuristic data augmentation model based on the state-of-the-art GP-WGAN network, which generates small and medium-sized synthetic defects which are heuristically allocated on the augmented samples and annotated. This algorithm is suitable for image object detection and segmentation and creates a sheer amount of high-quality synthetic images.
    \item We implement an enhanced Faster R-CNN as the automated object detection tool for defect detection and localization, using state-of-the-art Feature Pyramid Network (FPN) with Residual Network (ResNet).
    \item We analyze the performance of the algorithm with different settings and compare it with some other state-of-the art methods.
\end{itemize}

The rest of the paper is organized as follows. Section \ref{section:Methods} introduces the Faster R-CNN principle and the gradient-penalty GAN model. In section \ref{section:Proposed} the proposed data augmentation and the AVI model architecture are explained. Section \ref{section:Results} is devoted to illustrate the results of our experiments and the performance of the tested models. Finally, conclusions and future research paths are summarized in section \ref{section:Conclusion}.
\section{Methods and Materials}
\label{section:Methods}
% \subsection{ResNet}
\subsection{Faster R-CNN Principle}
In this section the basic idea of Faster R-CNN is introduced, being one of the cores of the proposed fault localization algorithm.

R-CNN, proposed by \cite{Girshick2014}, is intrinsically a Convolutional Neural Network (CNN) coupled with a region-proposal algorithm that suggests the object locations within an image. A selective search is designed to extract a fixed number of regions. The similar regions are merged together afterwards to obtain the candidate regions to be fed into the object detection algorithm. To cope with the slow runtime due to the comprehensive CNN feature extraction, \cite{Girshick2015} introduced Fast R-CNN, which employs a shared convolutional feature map as the output of the the CNN layer. This let the region of interests (RoI) to be extracted from these feature maps subsequently and faster.

By combining R-CNN and Fast R-CNN, \cite{Ren2017} proposed Faster R-CNN which demonstrated superior object detection capabilities in different domains. Like R-CNN and Fast R-CNN, Faster R-CNN divides the detection problem into two parts: (1) the region proposal stage, whose goal is to suggest the regions where objects may exist, and (2) the detection stage, where each selected region proposal is classified into different classes. In the first stage, the algorithm outputs the locations alongside with a binary object score which indicates if the region contains the corresponding object. In the detector stage, the algorithm generates class probabilities and refined region locations. By substituting the traditional region proposal stage with a CNN block, called Region Proposal Network (RPN), the network is able to share its layers with the object detector network, resulting in a reduced detection time.

Faster R-CNN architecture can be divided into three parts: the convolutional layer, which serves as a feature extractor, the RPN module, which tells the Fast R-CNN network where to look, and the Fast R-CNN network, which performs the classification and the bounding box estimation tasks.

\subsection{GP-GAN Principle}
In a typical GAN model there are two networks being trained simultaneously: (1) the discriminator, $D$, which is aimed to identify the real and false images as accurately as possible and to maximize the discriminant accuracy, and (2) the generator, $G$, which strives to deceive the discriminator. Essentially, the Nash equilibrium is obtained as follows:
\begin{equation}\label{eq:1}
\min_{G}\max_{D}\mathbb{E}_{x\sim \mathscr{P}_{r}}[\log(D(x))] + \mathbb{E}_{\tilde x \sim \mathscr{P}_{g}}[\log(1-D(\tilde x))]
\end{equation}
where $\tilde x = G(z)$ denotes a sample generated from random noise $z \sim  \mathscr{P}_{z}$, while $\mathscr{P}_{r}$, $\mathscr{P}_{g}$ and $\mathscr{P}_{z}$ indicate the data distribution of the real samples, generated samples and the random noise, respectively.
There are two general issues that traditional GAN models encounter during the training: first, when the discriminator becomes more accurate and gets higher performances, there's a higher probability that the generator gradient vanishes. Second, minimizing the loss function of the generator is equivalent to minimizing an unreasonable distance measure, leading to gradient instability.

In order to improve the stability of the GAN learning and cope with the mentioned issues, Arjovsky et al. \cite{Arjovsky2017} proposed a Wasserstein GAN, which employs the \emph{Earth-Mover} distance, $\mathcal{W}(\mathscr{P}_{r},\mathscr{P}_{g})$, instead of the Jensen-Shannon divergence. The \emph{Earth-Mover} distance can be interpreted as the minimum cost of transporting mass to transform the distribution $\mathscr{P}_{r}$ into the distribution $\mathscr{P}_{g}$.
\begin{equation}\label{eq:2}
\mathcal{W}(\mathscr{P}_{r},\mathscr{P}_{g})=\sup_{\|\mathcal{L}\|_L \leq 1}\mathbb{E}_{x\sim \mathscr{P}_{r}}[\mathcal{L}(x)] - \mathbb{E}_{x\sim \mathscr{P}_{g}}[\mathcal{L}(x)]
\end{equation}
where the supremum is over all the 1-Lipschitz functions $\mathcal{L}:\mathcal{X} \to \mathbb{R}$. The optimal set of 1-Lipschitz functions, $\mathcal{L}^*$, yield the optimal solutions of the following maximization problem:
\begin{equation}\label{eq:3}
\max_{\|\mathcal{L}\|_L \leq 1} \mathbb{E}_{\tilde{x}\sim \mathscr{P}_{r}}[\mathcal{L}(\tilde{x})] - \mathbb{E}_{x\sim \mathscr{P}_{g}}[\mathcal{L}(x)]
\end{equation}
Using the Kantorovich-Rubinstein duality, the WGAN objective function can be determined as:
\begin{equation}\label{eq:4}
\min_{G}\max_{D\in \mathcal{L}^*}\mathbb{E}_{x\sim \mathscr{P}_{r}}[D(x)] - \mathbb{E}_{\tilde x \sim \mathscr{P}_{g}}[D(\tilde x)]
\end{equation}
Let $\delta\sim U [0,1]$ be a random number to linearly interpolate between $x$ and $\tilde{x}$:
\begin{equation}\label{eq:5}
\hat{x}=\delta x + (1-\delta)\tilde{x}
\end{equation}
Consequently, the loss function of the discriminator, $\mathscr{L}_D$, can be determined as follows:

\begin{equation}\label{eq:6}
\mathscr{L}_D=\mathbb{E}_{\tilde{x}\sim \mathscr{P}_{g}}[D(\tilde{x})] - \mathbb{E}_{x\sim \mathscr{P}_{r}}[D(x)] + \lambda \mathbb{E}_{\hat{x}\sim \mathscr{P}_{z}}[(\|\nabla_{\hat{x}}D(\hat{x})\|_2 -1)^2]    
\end{equation}

where $\lambda$ stands for the gradient penalty coefficient. The last part of Eq.~\ref{eq:6}, $\lambda \mathbb{E}_{\hat{x}\sim \mathscr{P}_{z}}[(\|\nabla_{\hat{x}}D(\hat{x})\|_2 -1)^2]$, denotes the gradient penalty. Table~\ref{table:GPWGAN} shows the pseudo-code of GP-WGAN.

\begin{table}[t]
 \caption{GP-WGAN pseudo-code}
 \label{table:GPWGAN}
 \begin{tabular}{|p{12cm}|}
 \hline
  \begin{algorithm}[H]
  \KwInput{$\lambda, n_{critic}, m, \alpha, \beta_{1}, \beta_{2}, \omega_{0}, \theta_{0}$}
    \While{$\theta$ is not converged}{
    \For{$t \leftarrow 1,...,n_{critic}$}{
    \For{$i \leftarrow 1,...,m$}{
    Sample from real dataset $x\sim \mathscr{P}_x$,\\ Generate noise samples $z\sim \mathscr{P}_z$,\\ Generate a random number $\delta\sim U [0,1] $ \\
    $\tilde x \leftarrow G_{\theta}(z)$ \\
    $\hat{x} \leftarrow \epsilon x+(1-\delta)\tilde x$ \\
    $\mathscr{L}^{(i)} \leftarrow D_{\omega}(\tilde x) -D_{\omega}(x)+\lambda(\| \nabla_{\tilde x}D_{\omega}(\hat x) \|_{2}-1)^{2}$
    }
    $\omega \leftarrow Adam(\nabla_{\omega}\frac{1}{m}\sum_{i=1}^{m}L^{i},\omega,\alpha,\beta_{1}, \beta_{2})$
    }
    Sample batch of $m$ noise samples $\{z^{i}\}^{m}_{i=1}\sim \mathscr{P}_z$ \\
    $\theta \leftarrow Adam(\nabla_{\theta}\frac{1}{m}\sum_{i=1}^{m}-D_{\omega}(G_{\theta}(z)),\omega,\alpha,\beta_{1}, \beta_{2})$
    }
  \end{algorithm}
  \\
  \hline
  \end{tabular}
\end{table}

\section{The Proposed Framework}
\label{section:Proposed}

\subsection{Data Augmentation}
\begin{figure}
    \centering
    \includegraphics[width=12cm]{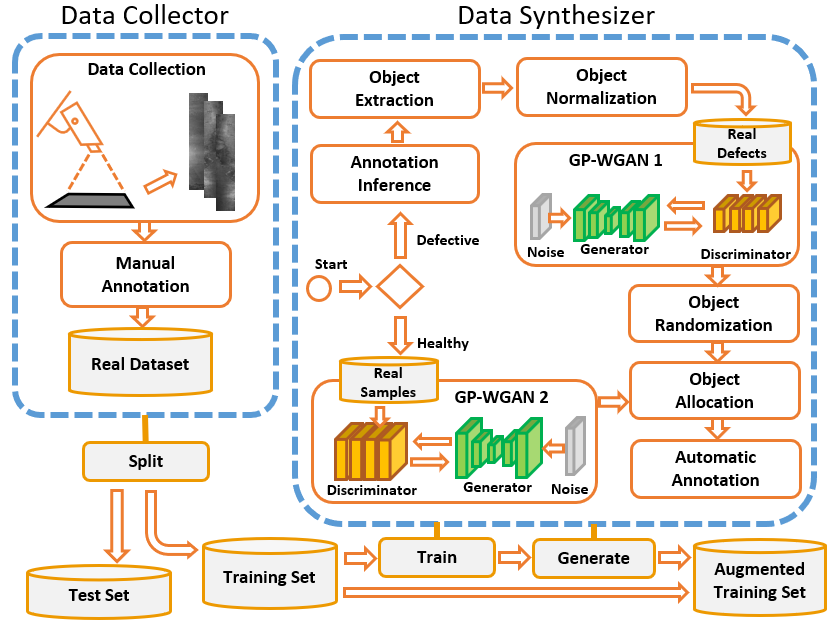}
    \caption{The proposed data augmentation framework based on GP-WGAN}
    \label{fig:augmentation mdoel}
\end{figure}

In this paper, in order to cope with insufficient number of samples for some fault types, we propose a framework which generates new high-quality synthetic samples with multiple defect types and annotates them subsequently. The sequence of this generative algorithm is illustrated in Fig.~\ref{fig:augmentation mdoel}.

By feeding the preprocessed bounding boxes to the GP-WGAN model, the model generates high-quality defects which are suitable to be processed afterwards, and to be randomly allocated on the synthetic and real defect-free image beds.

\subsection{Fault Diagnosis System}
\begin{figure}
    \centering
    \includegraphics[width=12cm]{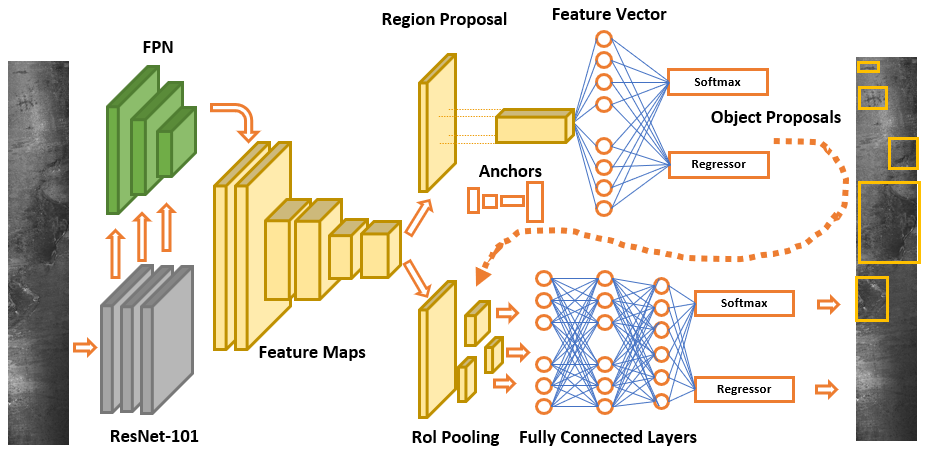}
    \caption{The architecture of FPN ResNet Faster R-CNN}
    \label{fig:faster R-CNN}
\end{figure}

The depth of the deep learning architecture is a key factor in achieving good results. Yet, when the network depth increases, the feature level correspondingly rises, which generates the gradient vanishing issue. To avoid that, we resort to ResNet-101 which uses the layers that are basically composed of a series of residual errors, degenerating the neural network model into a shallow network that solves the vanishing gradient problem. In our framework, the ResNet101 architecture aims at extracting the initial feature maps of the raw images, based on which the region proposal networks of the Faster R-CNN model can work more effectively. Furthermore, we use a Feature Pyramid Network (FPN) layer that utilizes a hierarchical feature layers and generates integrated feature maps that can push up the detection accuracy, more specifically on the small objects\cite{Ghiasi2019}. The overall structure of the proposed FPN ResNet-101 Faster R-CNN model is shown in Fig.~\ref{fig:faster R-CNN}. 

\section{Experimental Results and Discussions}
\label{section:Results}
All the experiments were run on python 3.8 (PyTorch 1.8) on an Ubuntu 18, with CUDA10.2, GPU of Nvidia Tesla K80, and 12GB of memory as well as a CPU of Intel(R) Xenon with 2.30GHz of frequency.

For the competing panel, we selected some state-of-the-art defect detection architectures which have been recently proposed in the literature and represented by Deconvolutional single shot detector (DSSD) \cite{Fu2017}, YOLO-v3 \cite{Redmon2018}, and Faster R-CNN \cite{Ren2017} with three different backbones: VGG16, FPN-ResNet-50, and FPN-ResNet-101. Running each network one time with the classic data augmentation strategy (e.g. by flipping, rotating and changing the image contrast), and the other time with the proposed data augmentation algorithm, the impact of the proposed GP-GAN-based technique on the results was examined. We also conducted a sensitivity analysis on the number of real samples ($m_r$) and the number of generated samples ($m_g$) to assess the detection ability of the proposed framework on the minority classes.
\subsection{The Performance Metrics}
To evaluate the performance of different algorithms, we used the mean Average Precision ($mAP$) metric, which is the average of the Average Precision($AP$) values of $\mathcal{C}$ different defect types. AP values are in fact the area under the ROC curve, determined using the following formula:
\begin{equation}\label{eq:7}
AP = \int_{0}^{1} p(r)\,dr = \sum_{k=1}^{M}P(k)\Delta r(k)
\end{equation}
Using Eq.~\ref{eq:7}, the $mAP$ values are calculated as follows:
\begin{equation}\label{eq:8}
mAP = \frac{\sum_{i=1}^{\mathcal{C}} AP_{i}}{\mathcal{C}}
\end{equation}

\subsection{Case Study 1}
For the first case study we analysed a steel surface defect dataset named NEU, containing six types of defects: scratches, rolled-in scale, inclusion, crazing, pitted surface and patches. For each defect type the dataset includes 300 images. In this experiment three-fold cross-validation was used. At each run, therefore, the test set comprised 100 images. In order to evaluate the performance of the model in imbalanced conditions, we randomly dropped out, from the training set, 25, 50, 75, 100 and 150 images related to the ``inclusion'' defect type.

\begin{figure}
    \centering
    \includegraphics[width=12cm]{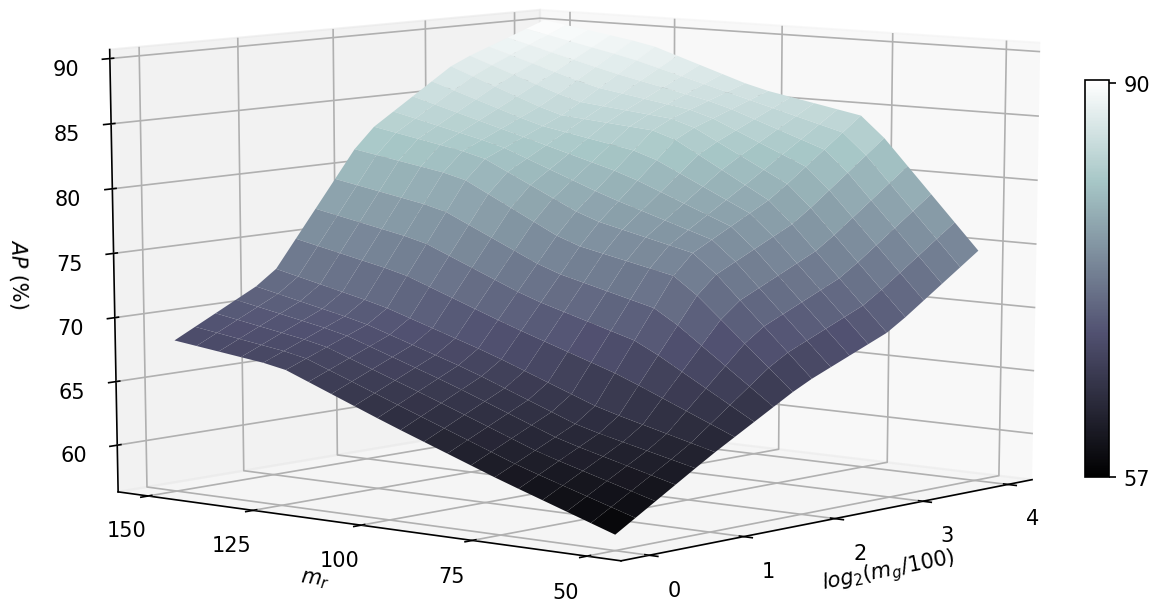}
    \caption{Sensitivity analysis of $AP_{inc.}$ on the NEU dataset}
    \label{fig:steel-map}
\end{figure}

Table \ref{table:steel} reports the overall $mAP$ values and the $AP_{inc.}$, for the minority class, ``inclusion'', when $m_r^{inc.}=50$. As it can be clearly seen in the table, the proposed data augmentation technique effectively improved the detection performance of all the five architectures. Specifically, it boosted the $AP$ values of DSSD-513 by 16\%, of YOLOv3 by 13\% and of different Faster R-CNN networks by more than 18\%. In this case, the proposed network reached  an $mAP$ equal to 80.9\%, significantly outperforming the other algorithms. Fig.~\ref{fig:steel-map} shows the sensitivity analysis of $AP$ for the class ``inclusion'' on 25 points with respect to different $m_r^{inc.}$ and $m_g$. When $m_r^{inc.}=250$ and $m_g=1600$, the proposed framework achieves an average precision of 90.2\% on the minority class.

\begin{table}[]
    \caption{Inspection results on the NEU dataset}
    \label{table:steel}
\begin{tabular}{|c|c|c|c|c|}
\cline{1-5}
Detector      & Backbone & Aug. & $mAP$  & $AP_{inc.}$ \\ \cline{1-5}
DSSD-513      & -        & classic         & 59.8 & 51.9   \\
YOLOv3       & DarkNet  & classic     & 59.2 & 49.3   \\
Faster RCNN         & VGG16    & classic         & 59.0 & 45.7  \\
Faster RCNN     & FPN ResNet50 & classic        & 68.5 & 48.1  \\
Faster RCNN     & FPN ResNet101 & classic    & 72.8 & 54.5  \\
DSSD-513      & -          & ours  & 66.1 & 67.7  \\
YOLOv3       & DarkNet    & ours  & 66.3 & 62.1  \\
Faster RCNN         & VGG16     & ours    & 74.4 & 65.3  \\
Faster RCNN     & FPN ResNet50 & ours    & 78.1 & 67.6  \\
\textbf{Faster RCNN} & \textbf{FPN ResNet101} & \textbf{ours}   & \textbf{80.9} & \textbf{72.5} \\
\cline{1-5}
\end{tabular}
\end{table}

\subsection{Case Study 2}
In the second case study we used a magnetic tile surface defect dataset \cite{Huang2020a} of 1344 images, composed of five types of defects: blowhole, crack, fray, break and uneven. We translated the segmentation annotations to the bounding box annotations making it compatible with the object detection algorithms. Three-fold cross-validation was applied also in this case, to have even-handed results. In order to evaluate the performance of the model in different imbalanced conditions, we randomly selected 50, 75, 125, 150 and 175 samples from the ``fray'' class to construct the dataset.

\begin{figure}
    \centering
    \includegraphics[width=12cm]{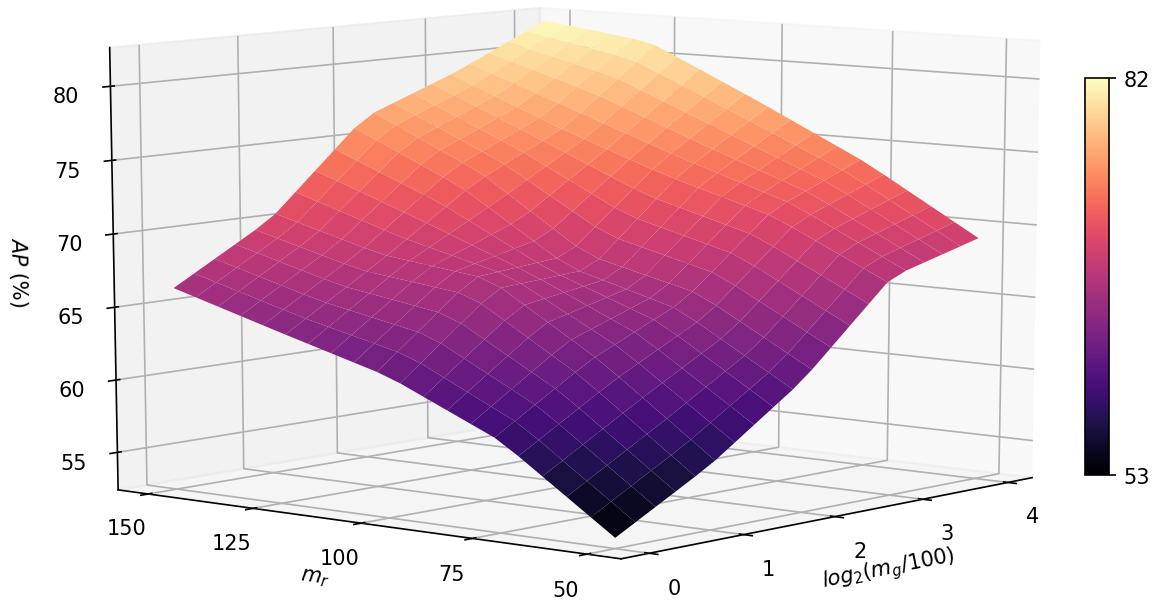}
    \caption{Sensitivity analysis of $AP_{fray}$ on the magnetic tile dataset}
    \label{fig:magnet}
\end{figure}

Table \ref{table:magnet} compares the performance of different models in terms of $mAP$ values and $AP_{fray}$ corresponding to the minority class (fray), when $m_r^{fray}=50$. As for the first case study, this comparison demonstrates the effectiveness of the proposed data augmentation technique on the performance of the detectors. Without any augmentation, Faster R-CNN with FPN backbones and the DSSD-513 dominate the other two detectors. In this situation, compared to the others, DSSD-513 gets the highest $AP$ value on the minority class. However, with the proposed data augmentation, it is Faster R-CNN with FPN ResNet-101 that outperforms the others in terms of both $mAP$ and $AP$ for the ``fray'' class. Fig.~\ref{fig:magnet} illustrates the performance of the proposed model on the minority class with different $m_r^{fray}$ and $m_g$ values.

\begin{table}[]
    \caption{Inspection results on the magnet tile dataset}
    \label{table:magnet}
\begin{tabular}{|c|c|c|c|c|}
\cline{1-5}
Detector      & Backbone & Aug. & $mAP$  & $AP_{fray}$ \\ \cline{1-5}
DSSD-513      & -        & classic         & 59.4 &	44.5\\
YOLOv3       & DarkNet  & classic         & 49.5 &	37.8\\
Faster RCNN         & VGG16    & classic         & 54.3 &	35.4\\
Faster RCNN     & FPN ResNet50 & classic         & 60  &	39.7\\
Faster RCNN     & FPN ResNet101 & classic         & 62.8 &	41.2\\
DSSD-513      & -          & ours  & 69.7 &	64.2\\
YOLOv3       & DarkNet    & ours  & 66.3 &	53.1\\
Faster RCNN         & VGG16     & ours    & 71.8 &	55.8\\
Faster RCNN     & FPN ResNet50 & ours    & 73	& 62.5\\
\textbf{Faster RCNN} & \textbf{FPN ResNet101} & \textbf{ours} & \textbf{76.9} & \textbf{69.1} \\

\cline{1-5}
\end{tabular}
\end{table}

\begin{figure}
    \centering
    \includegraphics[width=12cm]{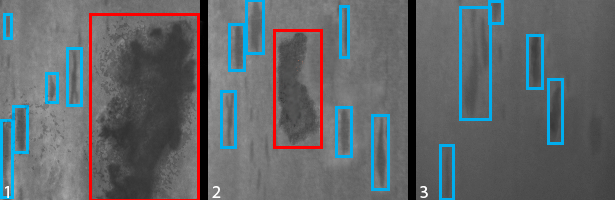}
    \caption{Three fake images with different synthetic defects}
    \label{fig:steel-fake}
\end{figure}

\begin{figure}
    \centering
    \includegraphics[width=12cm]{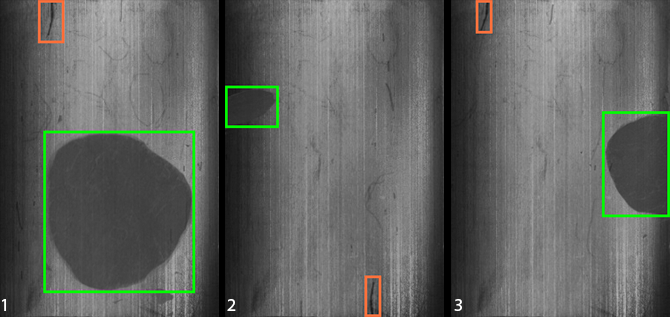}
    \caption{Three fake images with random defect allocations on the same defect-free image}
    \label{fig:magnet-fake}
\end{figure}

Fig.~\ref{fig:steel-fake} and Fig.~\ref{fig:magnet-fake} depict some of the fake defective samples generated by the proposed model on the first and second case studies, respectively.

\section{Conclusions and Future Directions}
\label{section:Conclusion}
Efficient AVI systems are one of the highly demanded parts
in the realization of Industry 4.0 and smart manufacturing systems. Deep learning-based object detection algorithms require a plethora of samples for their training phase, whereas the real-world datasets lack a sufficient number of annotated defects to be used. This problem hinders the proposed algorithms from attaining satisfactory results. The main contribution of this paper is to present a generative algorithm based on GP-WGAN architecture, which generates high-quality defect images and defect-free images. The algorithm then blends them and synthesizes new images with their corresponding annotations. In this paper, we also used a powerful object detection architecture, named Faster R-CNN with a state of the art backbone, FPN-ResNet-101, which can extract the
complex features that the object detector needs and enhance its accuracy. The comparison results on two industrial datasets applying a range of imbalanced severities indicate the effectiveness of the proposed augmentation technique and of the detection algorithm. Therefore, the industrial practitioners with a limited number of annotated samples or highly imbalanced data can benefit from the implementation of our data generation model. For future work, we will investigate the opportunities to reduce the training time and the inference time of the detection algorithm to work with higher frame per-second (FPS) rates.

\bibliography{AVI}

\end{document}